\documentclass[10pt,twocolumn,letterpaper]{article}

\usepackage[pagenumbers]{cvpr} 
\usepackage[dvipsnames]{xcolor}

\definecolor{cvprblue}{rgb}{0.21,0.49,0.74}
\usepackage[pagebackref,breaklinks,colorlinks,citecolor=cvprblue]{hyperref}

\usepackage[numbers]{natbib}
\newcommand{\modelname}{\textcolor{black}{ROG}\xspace}

\definecolor{DeltaColor}{rgb}{0.039,0.73,0.71}
\definecolor{SigmaColor}{rgb}{0.98,0.45,0.0}
\definecolor{AlphaColor}{rgb}{0,0,0.8}
\definecolor{BetaColor}{rgb}{0.8,0,0.8}
\definecolor{GammaColor}{rgb}{0.514,0.34,0.224}
\definecolor{EpsilonColor}{rgb}{0.353,0.725,0.906}
\definecolor{PurpleColor}{rgb}{0.5,0,0.7}
\definecolor{OrangeColor}{rgb}{0.914,0.541,0.141}
\definecolor{GreenColor}{rgb}{0.137,0.573,0.565}
\definecolor{RedColor}{rgb}{0.949,0.275, 0.224}
\definecolor{LightCyan}{rgb}{0.88,1,1}
\definecolor{Gray}{gray}{0.3}
\definecolor{Strawberry}{rgb}{1,0.26,0.64}

\definecolor{BetaColor}{rgb}{0.8,0,0.8}

\definecolor{LightCyan}{rgb}{0.88,1,1}
\definecolor{lightgray}{rgb}{0.9,0.9,0.9}

\definecolor{GreenColor}{rgb}{0.137,0.573,0.565}

\renewcommand{\paragraph}[1]{\medskip\noindent\textbf{#1}\ \ }

\makeatletter
\newcommand*\bigcdot{\mathpalette\bigcdot@{.5}}
\newcommand*\bigcdot@[2]{\mathbin{\vcenter{\hbox{\scalebox{#2}{$\m@th#1\bullet$}}}}}
\makeatother

\usepackage{color}
\usepackage{amsthm}
\usepackage{wrapfig}
\usepackage{array}
\usepackage{newlfont} %
\usepackage{textcomp}
\usepackage{multirow}
\usepackage{hhline}
\usepackage{tabularx}
\usepackage{algorithmic}
\usepackage{microtype}
\usepackage{mathtools}
\usepackage{booktabs} %
\usepackage[ruled]{algorithm2e} %
\usepackage{comment}
\usepackage{times}
\usepackage{epsfig}
\usepackage{graphicx}
\usepackage{amsmath}
\usepackage{cuted}
\usepackage{capt-of}
\usepackage{enumitem}
\usepackage{balance}
\usepackage[normalem]{ulem} 
\usepackage[tone, extra, safe]{tipa}
\usepackage{tipx}
\usepackage{amssymb}
\usepackage{enumitem}
\usepackage{svg}
\usepackage[dvipsnames]{xcolor}
\usepackage{arydshln}
\usepackage[toc,page]{appendix}
\usepackage[accsupp]{axessibility}

\def\paperID{7005} 
\def\confName{CVPR}
\def\confYear{2025}

\title{Guiding Human-Object Interactions with Rich Geometry and Relations}

\author{
Mengqing Xue$^1$\footnotemark[1] \quad
Yifei Liu$^1$\footnotemark[1] \quad
Ling Guo$^1$\footnotemark[1] \quad
Shaoli Huang$^3$ \quad
Changxing Ding$^{1,2}$\footnotemark[2] \\ 
$^1$South China University of Technology \quad 
$^2$Pazhou Lab, Guangzhou \quad 
$^3$Tencent AI Lab \\
{\tt\small \{eemengqing, ft\_lyf, eeguoling\}@mail.scut.edu.cn, shaoli.huang@gmail.com, chxding@scut.edu.cn}
}

\begin{document}

\captionsetup[figure]{hypcap=false}

\newcommand{\teasercaption}{Our proposed \modelname begins by leveraging rich geometric information to construct an Interactive Distance Field (IDF), effectively capturing the relational dynamics of Human-Object Interactions (HOI). It then utilizes the learned IDF prior to refine the generated motion's IDF, guiding the motion generation process to produce movements that are both relation-aware and semantically aligned. For clarity, we simplify the visualization by displaying only four key points for each object.}

\twocolumn[{
    \renewcommand\twocolumn[1][]{#1}
    \vspace{-0.5cm}
    \maketitle
    \centering
    
    \begin{minipage}{1.00\textwidth}
        \centering
        \includegraphics[width=\textwidth,height=0.40\textheight]{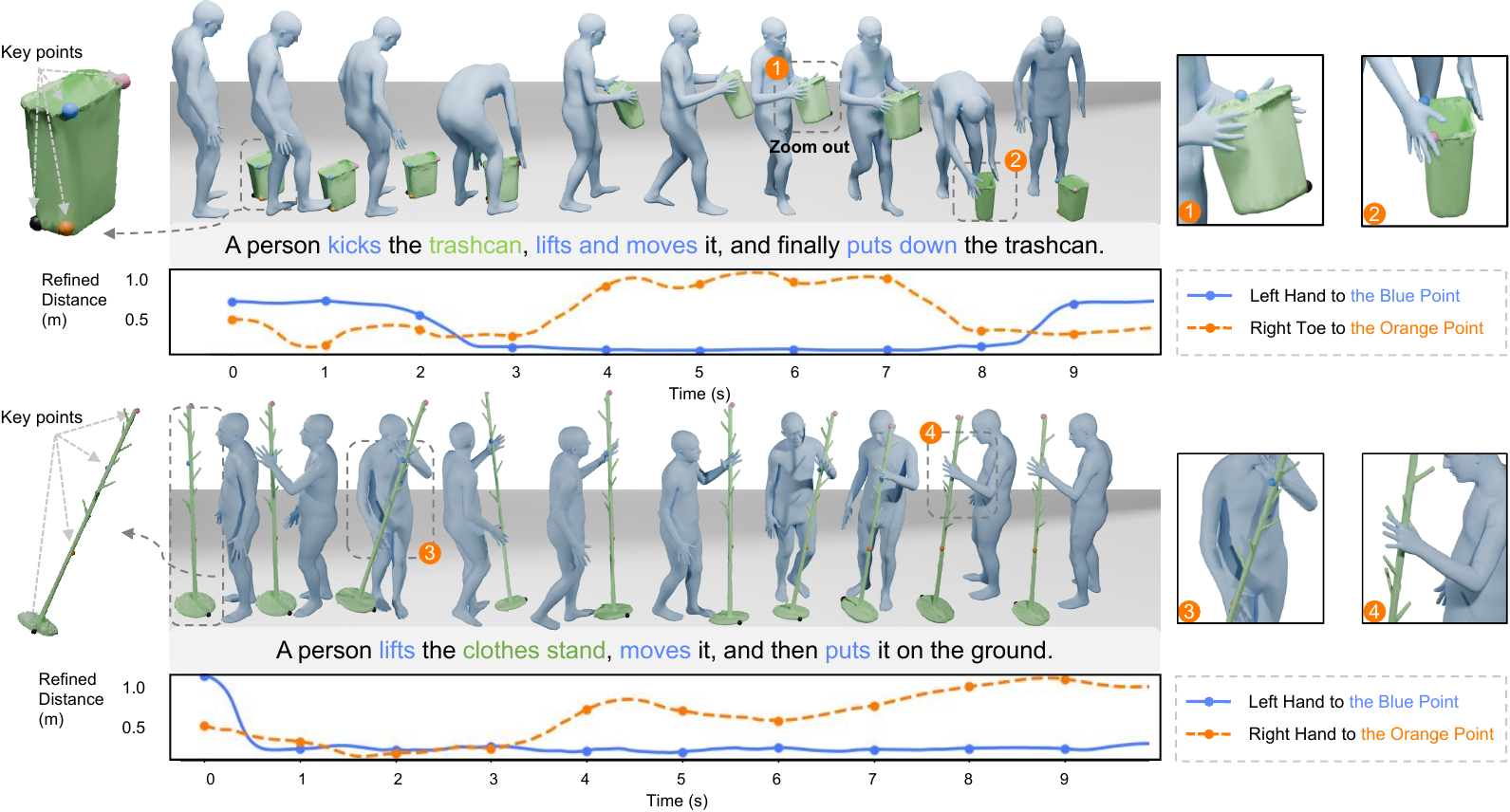}
    \end{minipage}
    \captionof{figure}{\teasercaption}
    \vspace{0.77cm}
    \label{fig:firstfigure}
}]

\footnotetext[1]{ The first three authors contribute equally.}
\footnotetext[2]{Corresponding author.}
\begin{abstract}
Human-object interaction (HOI) synthesis is crucial for creating immersive and realistic experiences for applications such as virtual reality. Existing methods often rely on simplified object representations, such as the object's centroid or the nearest point to a human, to achieve physically plausible motions. However, these approaches may overlook geometric complexity, resulting in suboptimal interaction fidelity.
To address this limitation, we introduce \textbf{\modelname}, a novel diffusion-based framework that models the spatiotemporal relationships inherent in HOIs with rich geometric detail. For efficient object representation, we select boundary-focused and fine-detail key points from the object mesh, ensuring a comprehensive depiction of the object's geometry. This representation is used to construct an interactive distance field (IDF), capturing the robust HOI dynamics.
Furthermore, we develop a diffusion-based relation model that integrates spatial and temporal attention mechanisms, enabling a better understanding of intricate HOI relationships. This relation model refines the generated motion's IDF, guiding the motion generation process to produce relation-aware and semantically aligned movements.
Experimental evaluations demonstrate that \modelname significantly outperforms state-of-the-art methods in the realism and semantic accuracy of synthesized HOIs. This paper’s code will be released at \href{https://lalalfhdh.github.io/rog_page/}{https://lalalfhdh.github.io/rog\_page/}.
\end{abstract}    
\section{Introduction}
Human-object interaction (HOI) synthesis is crucial for creating immersive and realistic experiences for applications such as virtual reality \cite{guzov2024interaction}, computer animation \cite{cao2024avatargo, starke2019neural, xie2023hierarchical}, and robotics \cite{bharadhwaj2023zero}. A vivid HOI sequence may consist of a series of actions involving an object, such as picking it up while walking, putting it down, and then adjusting its position with a foot, where the movements of the person and the object should be synchronized and the contact points must be precise. Achieving such lifelike HOIs requires generating natural body movements and an in-depth understanding of the spatiotemporal relationships between humans and the objects they manipulate.

Recently, the remarkable success of diffusion models \cite{ho2020denoising, nichol2021improved, ho2022classifier, dhariwal2021diffusion, liu2024sora} has spurred notable efforts to apply them in HOI generation \cite{zhang2024force, li2023object, cha2024text2hoi, song2024hoianimator, xu2023interdiff}. Some studies incorporate prior information into the motion generation model, such as fine-grained textual descriptions \cite{yang2024f}, physical forces \cite{zhang2024force}, hand joint positions \cite{li2023object}, and possible contact maps \cite{cha2024text2hoi, song2024hoianimator}. Meanwhile, a body of research have introduced guidance mechanisms to impose additional constraints on diffusion models, thereby achieving more physically plausible motions. For instance, Peng et al. \cite{peng2023hoi} and Li et al. \cite{li2023controllable} propose guidance mechanisms that enforce contact at selected keypoints during the denoising process. Xu et al. \cite{xu2023interdiff} and Diller et al. \cite{diller2024cg} introduce guidance by optimizing the distances between specific vertices at contact areas. Wu et al. \cite{wu2024thor} leverage the human motion as auxiliary guidance to correct the object's movement.

Despite these advances, the complex relationships between humans and objects remain underexplored due to certain challenges, resulting in suboptimal interaction fidelity. First, efficiently representing an object's geometry is difficult, hindering the accurate computation of spatiotemporal relationships between humans and objects such as spatial distance. Previous approaches \cite{wu2024thor, diller2024cg} often simplified object representation using its centroid or the nearest point relative to the human, neglecting the object's overall geometric complexity. Directly utilizing all the surface points on the object is also impractical due to the substantial increase in computational cost. Second, due to the high-dimensional and dynamic nature of HOIs, designing models that can learn these relationships is inherently challenging. Capturing subtle, context-dependent interactions requires sophisticated modeling techniques that effectively integrate spatial geometry with temporal dynamics.
 
To address these challenges, we introduce \textbf{\modelname}, a diffusion-based framework that comprehensively captures the spatiotemporal relationships in HOI through \textbf{R}ich ge\textbf{O}metric details, offering \textbf{G}uidance for the generation of more realistic interactions.
For the first problem, we propose an efficient object representation method utilizing Poisson disk sampling (PDS) \cite{yuksel2015sample}. Specifically, we begin by defining the smallest bounding box that encloses the object. Then, we select eight points on the object closest to the bounding box and apply PDS to obtain an additional sixteen points. Typically, boundary points are located at the object's extremities and corners, being essential for defining the object’s overall shape. Meanwile, PDS captures fine surface details and subtle geometric variations that may be typically overlooked. Combining boundary-focused and PDS points ensures a holistic representation of the object's geometry. Using this representation, we construct a three-dimensional (3D) matrix that measures the distance between human joints and the object keypoints over time, termed the interactive distance field (IDF). We use IDF as an additional objective function to improve the motion generation model's understanding of HOI dynamics.

For the second problem, we develop a diffusion-based relation model that captures spatiotemporal relation in HOI by predicting the IDF matrix. More specifically, drawing inspiration from successful modeling of three-dimensional data in the domain of video generation \cite{ho2022video, lu2023vdt, gupta2025photorealistic, zhang2023controlvideo, li2023videogen}, we incorporate spatial and temporal self-attention into our relation model to capture local interactions and temporal dependencies. We then design a guidance process in which the relation model directs the motion generation model to produce relation-aware movements. Specifically, during the denoising process, we compute the IDF matrix based on the motion predicted by the motion generation model. The relation model then takes this matrix as input and produces a refined IDF matrix that, in turn, corrects the motion, ensuring that the generated motions adhere closely to the desired spatiotemporal relationships.

By integrating these core designs, our approach enables the generation of more realistic and semantic-aligned human-object interactions. We quantitatively evaluate the realism and semantic correctness of our synthesized motion compared to ground truth and state-of-the-art methods on the FullBodyManipulation dataset \cite{li2023object}. Experimental results demonstrate that our model surpasses state-of-the-art methods both in qualitative and quantitative terms, advancing the field of human-object interaction synthesis.

\section{Related Work}
Existing approaches to enhancing HOI motion generation quality can be categorized into two paradigms, i.e., prior-based strategies and post-optimization-based strategies. The prior-based strategies focus on incorporating prior information into the motion generation model. Various priors can be utilized, including textual descriptions of body part movements \cite{yang2024f, ghosh2022imos}, physical force \cite{zhang2024force}, hand joint position \cite{li2023object}, arm movements \cite{ghosh2022imos}, keyframe poses \cite{li2024task, wu2022saga}, and possible contact maps \cite{cha2024text2hoi, song2024hoianimator}. However, owing to the lack of direct supervision on the spatial relation between human and object, the generated results are likely to present artifacts, e.g., inconsistent contact, interpenetration, and floating.

The post-optimization-based strategies typically leverage a meticulously designed subsequent process combined with a specific human-object relation representation to optimize the generated coarse motion. Based on the human-object relations representation, existing works approaches can be classified into (i) contact-based methods \cite{peng2023hoi, li2023controllable}, (ii) motion-based methods \cite{wu2024thor}, and (iii) distance-based methods \cite{xu2023interdiff, diller2024cg, kulkarni2023nifty}. The contact-based methods typically select a fixed set of body joints and use a generation module to predict the contact probability of each joint and the corresponding vertices on the object. In contrast, the motion-based methods focus on the differences in translation and rotation between human joints and the object. They then incorporate this information to correct the object motion. The distance-based methods, on the other hand, compute the distances between specific vertices on the human and the object to supervise or adjust the generated coarse motions. Additionally, most of the works mentioned above apply various forms of guidance during inference \cite{peng2023hoi, li2023controllable, wu2024human, xu2023interdiff, diller2024cg}, such as distance-based constraints and spatial alignment techniques. These guidance methods help refine the output motion distribution, bringing it closer to the desired interaction distances. While human-object relation representations are addressed in these works, they primarily rely on a single view, i.e., the human joints and their nearest object vertices, without incorporating the mutual spatial relations between the human and object. This may affect the model's ability to capture complex HOIs with fine details, leading to suboptimal generation results.

To address this, our method samples and tracks vertices on both the human and the object to compute a mutual interaction distance field. Furthermore, we introduce a video diffusion transformer (VDT) module to effectively integrate distance information into generation through guidance.

\section{Method}

    \paragraph{Overview.}
Given an object mesh, a human body skeleton, and a text prompt, our diffusion-based framework generates human-object interactions which are semantically-aligned with the text prompt. During each denoising step, the motion generation model first produces initial movements for both the human and the object. Subsequently, the relation model takes the Interactive Distance Field (IDF) derived from these initial movements as input and outputs a refined IDF. This refined IDF serves as guidance to enhance the quality of the initial movements.

To provide a comprehensive overview of our framework, we begin with a preliminary section that briefly introduces the data representation and our motion generation model, which is based on the Motion Diffusion Model \cite{tevet2022human} (Section \ref{sec:preliminary}). Following this, we introduce our Interactive Distance Field (IDF) and explain the process for computing the IDF (Section \ref{sec:idf}). Finally, we describe our relation model and the guidance mechanism in detail (Section \ref{sec:relation_guidance}).

\begin{figure*}[t!]
    \begin{center}
    \includegraphics[width=1.0\linewidth,height=7.5cm]{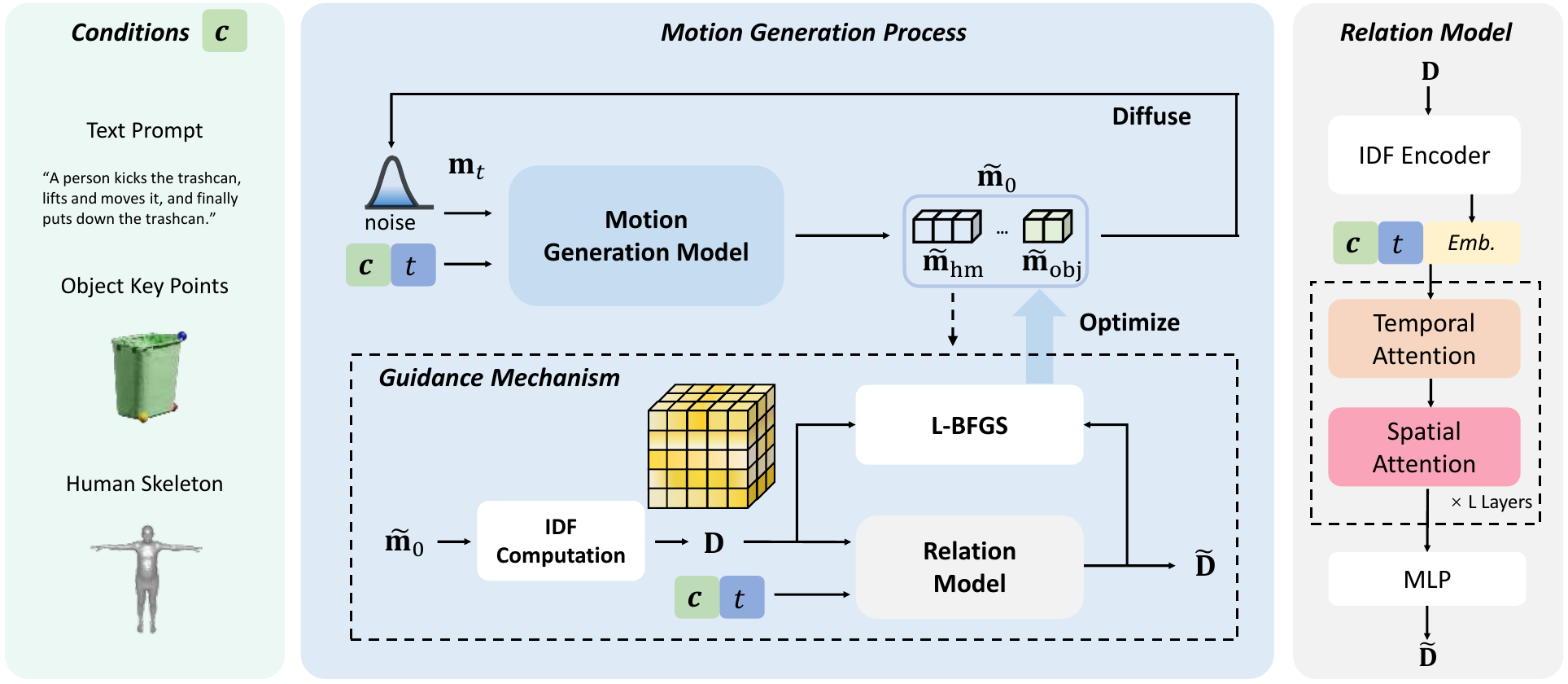}
    \end{center}

    \caption{Overview of \modelname.
    Given an object, \modelname first extracts key points that comprehensively represent the object's geometry. These object key points, along with human key points, a text prompt, and the diffusion step $t$, are then input into \modelname to generate human-object interactions that are semantically aligned with the text prompt. During each denoising step, the motion generation model initially produces movements $\tilde{\mathbf{m}}_0 = \{\tilde{\mathbf{m}}_\text{hm}, \tilde{\mathbf{m}}_\text{obj}\}$ for both the human and the object. The relation model then uses the Interactive Distance Field (IDF) $\mathbf{D}$ derived from these initial movements to output a refined IDF $\tilde{\mathbf{D}}$. This refined IDF guides the enhancement of the initial movements, improving their quality.
    }
    \label{fig:overview}

\end{figure*}

\subsection{Preliminary}
\label{sec:preliminary}

    \paragraph{Data representation.}
Our framework integrates multiple data modalities, including text, objects, human skeletons, and motion. Text embeddings are encoded using a pretrained CLIP model \cite{radford2021learning} and are represented as \( \mathbf{f}_{\text{text}} \in \mathbb{R}^{512} \). Objects are modeled by their mesh vertices \( \mathbf{V} = \{ \mathbf{v}_i \}_{i=1}^K \), where each vertex \( \mathbf{v}_i \in \mathbb{R}^3 \) denotes its position in 3D space, and \( K \) is the total number of vertices. key points sampled on the object mesh are denoted by $\mathbf{P} = \{\mathbf{p}_{1}, \mathbf{p}_{2}, \ldots, \mathbf{p}_{24}\}$. The human skeleton is represented by $\mathbf{Q} = \{\mathbf{q}_{1}, \mathbf{q}_{2}, \ldots, \mathbf{q}_{24}\}$, indicating 24 key points corresponding to human joints. Motion data is expressed as \( \mathbf{m} = \{ \mathbf{m}_{\text{hm}}, \mathbf{m}_{\text{obj}} \} \in \mathbb{R}^{N \times 288} \), where \( N \) denotes the number of frames in the sequence. \( \mathbf{m}_{\text{hm}} \in \mathbb{R}^{N \times 204} \) captures the sequence of human motions, including joint positions and 6D continuous rotations \cite{zhou2019continuity}, and \( \mathbf{m}_{\text{obj}} \in \mathbb{R}^{N \times 84} \) captures the sequence of object motions, encompassing object transitions, rotations, and key point positions.

    \paragraph{Motion Generation Model.}
Our motion generation model is based on the Motion Diffusion Model (MDM) \cite{tevet2022human}, which learns to reconstruct motion data from noise through a series of time steps. Starting with an initial motion sequence \( \mathbf{m}_0 \) sampled from the data distribution, we apply the diffusion process to generate a sequence of progressively noisier data \( \{ \mathbf{m}_t \}_{t=1}^T \), where \( T \) denotes the total number of diffusion steps:
\begin{equation}
q\left( \mathbf{m}_t \mid \mathbf{m}_{t-1} \right) = \mathcal{N}\left( \sqrt{\alpha_t} \, \mathbf{m}_{t-1},\ (1 - \alpha_t) \mathbf{I} \right),
\end{equation}
where \( \alpha_t \in (0, 1) \) is a time-dependent variance schedule parameter, \( \mathcal{N}(0, \mathbf{I}) \) represents a standard normal distribution, and \( \mathbf{I} \) denotes the identity matrix. As \( t \) increases, the distribution of \( \mathbf{m}_T \) progressively approaches \( \mathcal{N}(0, \mathbf{I}) \).

The motion generation model \( G \) is trained to approximate the reverse diffusion process, modeling the conditional distribution \( p(\mathbf{m}_0 \mid \mathbf{c}) \). This is achieved by iteratively denoising \( \mathbf{m}_T \) to reconstruct the original motion sequence. The training objective is to minimize the following reconstruction loss:
\begin{align}
\mathcal{L}_\text{rec} &= \mathbb{E}_{\mathbf{m}_0, t} \left\| \mathbf{m}_0 - \tilde{\mathbf{m}}_0 \right\|_2^2, \\
\tilde{\mathbf{m}}_0 &= \{\tilde{\mathbf{m}}_\text{hm}, \tilde{\mathbf{m}}_\text{obj}\} = G\left( \mathbf{m}_t, t, \mathbf{c} \right),
\end{align}
where \( \tilde{\mathbf{m}}_0 \) is the motion predicted by the model \( G \), comprising the predicted human motion \( \tilde{\mathbf{m}}_\text{hm} \) and the predicted object motion \( \tilde{\mathbf{m}}_\text{obj} \). \( \mathbf{c} = \{ \mathbf{f}_\text{text}, \mathbf{Q}, \mathbf{P} \} \) represents the conditioning information including the text embedding \( \mathbf{f}_\text{text} \), object key points in the default pose \( \mathbf{Q} \), and human key points in the default pose \( \mathbf{P} \).

\subsection{Interactive Distance Field}
\label{sec:idf}

Generating realistic Human-Object Interactions (HOI) requires accurately modeling the spatial relationships between humans and objects. Existing approaches \cite{li2023controllable, wu2024thor, peng2023hoi} have significantly advanced this area by representing object translations using object centroids \cite{wu2024thor} or leveraging contact points between human and object surfaces \cite{diller2024cg}. While these methods effectively capture certain aspects of spatial dynamics, they may not fully encompass the comprehensive spatial relationships inherent in complex interactions. Building upon these foundations, we introduce the Interactive Distance Field (IDF), a novel approach that calculates the spatial proximity between the human and object through a comprehensive matrix representation.

A key aspect of establishing the IDF matrix is the efficient representation of the object's shape and structure. To achieve this, we sample key points on the object's surface \(\mathbf{V}\) to capture its geometry, including both its rough outline and detailed features. First, we establish the object's Axis-Aligned Bounding Box (AABB) \cite{schneider2002geometric}, the smallest cuboid that fully encloses the object based on its extreme coordinates in 3D space. We then identify the surface points closest to the AABB vertices to outline the object's boundaries. To achieve a more refined and comprehensive representation of the object's shape, we employ Poisson Disk Sampling \cite{yuksel2015sample}, a probabilistic method that enforces a minimum distance constraint between points to ensure uniform distribution across the object's surface \(\mathbf{V}\). Ultimately, we sample a total of twenty-four key points, denoted as \(\mathbf{P} = \{\mathbf{p}_{1}, \mathbf{p}_{2}, \ldots, \mathbf{p}_{24}\}\). Each point represents a significant surface feature on the 3D object mesh, providing a representative and precisely located set of points for detailed spatial analysis.

Parallel to object feature identification, we select twenty-four skeletal key points from the SMPL-X \cite{smpl, smplx, smplify} human model, represented as \(\mathbf{Q} = \{\mathbf{q}_{1}, \mathbf{q}_{2}, \ldots, \mathbf{q}_{24}\}\). These key points include major skeletal joints such as the spine, elbows, pelvis, and knees, as well as the palm centers of both hands, capturing the full spectrum of human motion dynamics essential for realistic HOI.

Building upon the identified object and human key points, we compute the distance matrix between these key points during the HOI, referred to as the Interactive Distance Field (IDF) matrix. Specifically, for a sequence of HOI motions spanning \(N\) frames, we construct the IDF matrix \(\mathbf{D}(\{\mathbf{Q}_n\}^N_{n=1}, \{\mathbf{P}_n\}^N_{n=1}) = [\mathbf{D}_{i,j,n}] \in \mathbb{R}^{24 \times 24 \times N}\). Each element \(\mathbf{D}_{i,j,n}\) represents the Euclidean distance between the \(i\)-th human joint and the \(j\)-th object point at frame \(n\):
\begin{equation}
    \mathbf{D}_{i,j,n} = \left\| \mathbf{q}_{i,n} - \mathbf{p}_{j,n} \right\|_2^2,
\end{equation}
where \(\mathbf{q}_{i,n}\) and \(\mathbf{p}_{j,n}\) denote the \(i\)-th human joint and the \(j\)-th object point at frame \(n\), respectively. By focusing on both human joint coordinates and salient object surface features, the IDF provides a detailed and comprehensive spatial mapping of interaction dynamics. Additionally, the matrix representation of distances facilitates effective modeling of complex spatial relationships, addressing the limitations of centroid-based and contact point-based approaches.

    \paragraph{Interactive Distance Field Loss.}
After establishing the Interactive Distance Field (IDF), we leverage it to directly supervise the training of the motion generation model. This supervision enables the model to develop a deeper understanding of the spatial relationships between humans and objects, thereby enhancing the accuracy and realism of the generated human-object interactions. Specifically, the IDF loss quantifies the discrepancy between the predicted IDF matrix \( \mathbf{D}_\text{pr} \) and the ground truth IDF matrix \( \mathbf{D}_\text{gt} \) using the Mean Squared Error (MSE):
\begin{align}
    \mathcal{L}_\text{IDF} &= \mathbb{E}_{\mathbf{m}_0, t} \| \mathbf{D}_\text{pr} - \mathbf{D}_\text{gt} \|_2^2 , \\
    \mathbf{D}_\text{pr} &= D\left( \{\tilde{\mathbf{Q}}_n\}_{n=1}^N, \{\tilde{\mathbf{P}}_n\}_{n=1}^N \right), \\
    \mathbf{D}_\text{gt} &= D\left( \{\mathbf{Q}_n\}_{n=1}^N, \{\mathbf{P}_n\}_{n=1}^N \right),
\end{align}
where \( \{\tilde{\mathbf{Q}}_n\}_{n=1}^N \) and \( \{\tilde{\mathbf{P}}_n\}_{n=1}^N \) denote the predicted human joints and object key points derived from the predicted motion sequence \( \tilde{\mathbf{m}}_0 \). \( \{\mathbf{Q}_n\}_{n=1}^N \) and \( \{\mathbf{P}_n\}_{n=1}^N \) represent the ground truth human joints and object key points from the ground truth motion sequence \( \mathbf{m}_0 \). \( D(\cdot) \) is the function that computes the IDF matrix from the provided key points. 
By minimizing \( \mathcal{L}_\text{IDF} \), the motion generation model \( G \) is encouraged to produce motion sequences that closely align with the spatial dynamics captured in the ground truth, resulting in more realistic and coherent human-object interactions.

Overall, the loss function for the motion generation model is formulated as:
\begin{align}
    \mathcal{L}_\text{m} = \mathcal{L}_\text{rec} + \lambda_\text{IDF}\mathcal{L}_\text{IDF},
\end{align}
where $\lambda_{\text{IDF}}$ is set to 5.0, controlling the contribution of the IDF loss to the total loss.

\subsection{Relation Model and Guidance Mechanism}
\label{sec:relation_guidance}

Given the Interactive Distance Field (IDF) that captures the dynamics of human-object interactions, our next objective is to learn this dynamic prior from real data. By doing so, we can provide informed guidance during the motion generation process, ensuring that the synthesized interactions maintain realistic spatiotemporal relationships. To achieve this, we introduce two key components: the \textbf{Relation Model}, which captures the intricate spatiotemporal relationships inherent in human-object interactions, and the \textbf{Guidance Mechanism}, which utilizes the outputs of the Relation Model to refine and enhance the generated motions.

    \paragraph{Relation Model.}
To effectively capture the intricate spatiotemporal relationships in human-object interactions, our Relation Model integrates both spatial and temporal self-attention mechanisms, drawing inspiration from space-time self-attention techniques commonly utilized in the video generation domain \cite{lu2023vdt}. The primary objective of the Relation Model is to learn a dynamic prior from real data distributions, which can guide the motion generation process to produce realistic and semantically aligned interactions.

Given the Interactive Distance Field (IDF) matrix at diffusion step \( t \), denoted as \( \mathbf{D}_t \in \mathbb{R}^{24 \times 24 \times N} \), our Relation Model \( R \) aims to reconstruct the ground truth IDF matrix \( \mathbf{D}_0 \). As illustrated in Figure~\ref{fig:overview}, to reduce computational complexity while preserving essential spatial information, we first apply a linear transformation that downsamples the spatial dimensions from \( 24 \times 24 \) to \( 4 \times 4 \). This dimensionality reduction allows the model to process the IDF more efficiently without imposing any predefined partitioning based on body parts or object regions, thereby enabling the network to autonomously identify and prioritize the most salient spatial features relevant to the interactions.

The dimension-reduced IDF tensor is then processed through multiple transformer blocks \cite{transformer} that integrate spatial and temporal self-attention mechanisms. Within each transformer block, spatial self-attention is applied across the \( 4 \times 4 \) spatial grid, enabling the model to aggregate and model dependencies between different spatial regions of the human body and the object. Concurrently, temporal self-attention is applied along the temporal dimension \( N \), which represents the sequence of interaction frames. This combination of spatial and temporal self-attention facilitates a holistic understanding of the spatiotemporal dynamics inherent in human-object interactions, ensuring that the Relation Model captures both spatial dependencies and the temporal evolution of interactions. 

To train the Relation Model, we aim to reconstruct the ground truth IDF matrix \( \mathbf{D}_0 \) from the noisy input \( \mathbf{D}_t \) using the following loss function:
\begin{align}
    \mathcal{L}_\text{D} &= \mathbb{E}_{\mathbf{D}_0, t} \left\| \mathbf{D}_0 - \tilde{\mathbf{D}}_0 \right\|_2^2, \\
    \tilde{\mathbf{D}}_0 &= R\left( \mathbf{D}_t, t, \mathbf{c} \right),
\end{align}
where \( \mathbf{c} = \{ \mathbf{f}_\text{text}, \mathbf{Q}, \mathbf{P} \} \) represents the condition information that is consistent with Eq.~3, and $R\left(\cdot \right)$ denotes our Relation Model.

    \paragraph{Guidance Mechanism.}
The motion generation model could produce unrealistic motions, exhibiting implausible contacts and dynamics, which are subsequently reflected in the Interactive Distance Field (IDF) during inference. To address this issue, we leverage the previously learned IDF prior to optimize the generated motion's IDF in the model inference stage. By propagating gradients from the IDF prior to the generated motion, we refine the motion to better align with realistic interaction dynamics.

At the denoising step \( t \) in the motion generation process, the motion generation model \( G \) predicts an initial motion sequence \( \tilde{\mathbf{m}}_0 \). We perform forward kinematics on \( \tilde{\mathbf{m}}_0 \) to obtain the corresponding human joints and object key points, which are then used to compute the IDF matrix \( \mathbf{D} \). This IDF matrix \( \mathbf{D} \) is subsequently fed into the Relation Model \( R \) to produce a refined IDF matrix \( \mathbf{\tilde{D}} \). The guidance loss is defined as:
\begin{equation}
L_{\text{guidance}} = \| \mathbf{D} - \mathbf{\tilde{D}} \|_2^2.
\label{eq:guidance}
\end{equation}
To minimize \( L_{\text{guidance}} \), we optimize the predicted motion sequence \( \tilde{\mathbf{m}}_0 \) using the second-order L-BFGS optimizer \cite{liu1989limited}, following the methodology proposed by Wang et al. \cite{wang2023intercontrol}. This optimization encourages the motion generation model to produce sequences that closely align with the realistic spatial dynamics captured in the ground truth IDF.

Realizing that motions generated in the early timesteps are particularly noisy \cite{karunratanakul2023guided}, we apply the guidance mechanism selectively during the last ten timesteps. In these final stages, the L-BFGS optimizer iteratively refines \( \tilde{\mathbf{m}}_0 \), enhancing the realism and coherence of the resulting human-object interactions. This targeted application ensures computational efficiency while significantly improving the quality of the synthesized motions.

Overall, the guidance mechanism effectively bridges the gap between initial motion predictions and realistic interaction dynamics by utilizing the refined IDF matrix. This results in more accurate and visually plausible human-object interactions during model inference, addressing the limitations of previous centroid-based and contact point-based approaches.

\section{Experiments}

\subsection{Dateset and Settings}
    \paragraph{Dateset.} 
The FullBodyManipulation dataset \cite{li2023object} provides high-quality HOI data. It includes 10 hours of human and object motion data, with corresponding videos featuring 17 subjects interacting with 15 objects. Following CHOIS \cite{li2023controllable}'s split, our training set consists of data from 15 subjects, while the testing set includes data from the remaining two subjects. Additionally, articulated objects, such as vacuums and mops, have been excluded from the dataset.

    \paragraph{Metrics.}
Following existing works \cite{song2024hoianimator, diller2024cg, li2023controllable}, we employ widely recognized evaluation metrics for HOI synthesis results: \textbf{Fréchet Inception Distance} (FID), which quantifies the discrepancy between real and generated motion distributions; \textbf{R-Precision}, which measures the alignment between generated motions and their corresponding textual prompts; and \textbf{Diversity}, which captures the variation within the generated motions. Additionally, we incorporate the \textbf{Foot Sliding score} (FS) \cite{li2023controllable} to assess the extent of foot sliding in the motion sequences.

We further introduce three additiponal metrics: Contact Percentage ($C_{\%}$) \cite{li2023controllable}, Collision Percentage ($Coll_{\%}$) \cite{li2023object}, and Motion Deviation (MDev) \cite{fan2023arctic}. The \textbf{Contact Percentage} measures the proportion of frames where human-object contacts are detected, while the \textbf{Collision Percentage} indicates the proportion of frames where collisions happen between the human and object meshes. For both metrics, we adopt the same threshold values as those in \cite{li2023object,li2023controllable} to define contacts or collisions. The \textbf{Motion Deviation} (MDev) \cite{fan2023arctic} measures the difference in movement directions between the human hand and object over a window of time. Therefore, it reflects whether the body parts are in stable contact with the object, and it uses the same threshold value as the Contact Percentage for its calculations.

    \paragraph{Implementation Details.} 
In alignment with MDM \cite{tevet2022human}, we encode text prompts using the `ViTB/32' variant of the CLIP model \cite{radford2021learning}. Our Motion Generation Model employs an encoder composed of 8 transformer layers and operates with a batch size of 64. For the Relation Model, we adopt the configuration from VDT \cite{lu2023vdt}, which includes 1 input channel, 16 patches, 8 blocks, a hidden layer size of 384, and a batch size of 8.
Both models are optimized using the AdamW optimizer \cite{loshchilov2017decoupled} with a learning rate of \(1 \times 10^{-4}\) over a total of 1,000 diffusion steps. We apply the DDPM sampling \cite{ho2020denoising} and select diffusion timesteps to implement guidance using the L-BFGS algorithm \cite{liu1989limited}. Details of the guidance setup are provided in the appendix. All experiments were conducted on an NVIDIA RTX 4090 GPU. 

\begin{table*}
  \centering
  \begin{tabular}{lccccccccc} %
    \toprule
     \multirow{2}{*}{Methods}  & \multicolumn{3}{c}{R-precision $\uparrow$}  & \multirow{2}{*}{FID $\downarrow$}  & \multirow{2}{*}{Diversity $\rightarrow$} & \multirow{2}{*}{FS $\downarrow$} & \multirow{2}{*}{$C_{\%}$} & \multirow{2}{*}{$Coll_{\%}$} & \multirow{2}{*}{MDev $\downarrow$} \\ 
    \cmidrule(r){2-4}  
     & Top-1     & Top-2     & Top-3 \\
    \midrule
     {Real motions}  & 0.651  & 0.862 & 0.917 & 0.001  & 8.693 & 0.222 & 0.623   & 0.157    & 4.846\\
    {InterGen \citep{liang2024intergen}} & 0.490  & 0.619 & 0.685 & 19.038 & 12.113& 0.603 & 0.179 & \textbf{0.156} & 39.795\\
     {MDM \citep{tevet2022human}} & 0.495  &0.626  & 0.681 & 9.775  & 9.972 & \textbf{0.331} & 0.349 & 0.210 & 9.549\\
     {HOI-Diff \citep{peng2023hoi}}  &0.534 & 0.646 &0.722  &11.875 & 11.063 & 0.430 & 0.372 & 0.175  & 58.728\\
     {CHOIS \citep{li2023controllable}} & 0.630 & 0.776 & 0.844 & 5.227 & 9.106 & 0.425 & 0.444 & 0.208  & 13.408\\
     {\modelname (Ours)}    &  \textbf{0.706}  & \textbf{0.835}   & \textbf{0.902} & \textbf{5.119}  & \textbf{8.863}& 0.349 & \textbf{0.466} & 0.200 & \textbf{5.815} \\
  \bottomrule
  \end{tabular}
  \caption{Quantitative comparisons on the FullBodyManipulation dataset \cite{li2023object}. We evaluate our model (\modelname) by comparing it with four baseline models, as well as with real motions from the test set. The symbol `$\rightarrow$' means results closer to those of the real motions are considered better.}
  \label{sample-table1}
\end{table*}

\begin{figure*}[t!]
    \centering
    \includegraphics[width=0.98\linewidth]{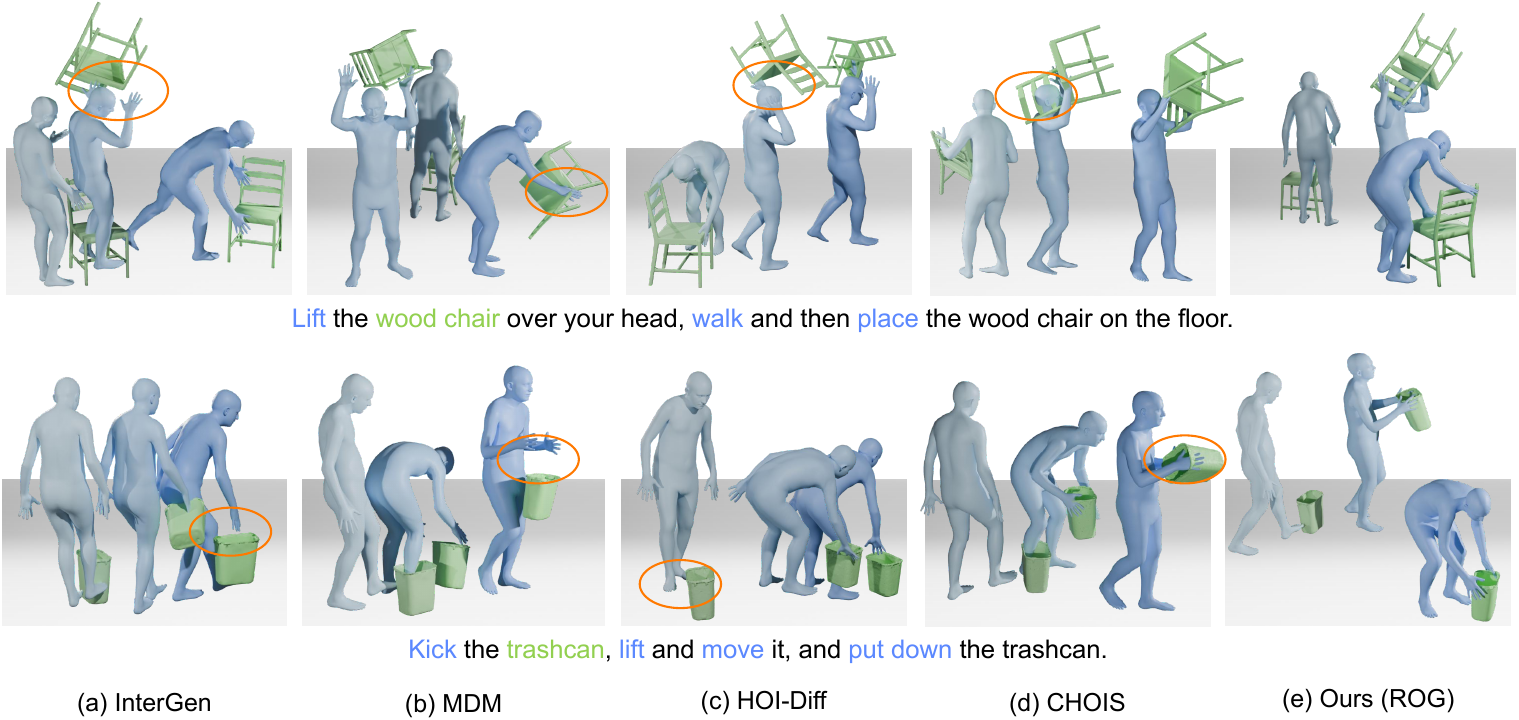}
    \caption{Qualitative comparisons. We use circles to highlight incorrect interactions, illustrating that our method can generate more realistic and physically plausible interactions that align with the given text.}
    \label{fig:compare}
\end{figure*}

    \paragraph{Baselines.} 
We make comparisons with multiple state-of-the-art open-source approaches, namely InterGen \citep{liang2024intergen}, MDM \citep{tevet2022human}, CHOIS \citep{li2023controllable}, and HOI-Diff \citep{peng2023hoi}, and adapt them to fit our task. Given InterGen's strong performance in synthesizing human-human interactions, we modify it by training cooperative networks for human-and-object motions instead of human-and-human motions. For MDM, a widely used method for text-to-motion generation, we expand its input and output dimensions to predict both object translations and rotations, enabling HOI synthesis. For CHOIS, which generates HOI by requiring additional control signals such as waypoints, we remove these signals and retrain the model. Finally, since HOI-Diff directly aligns with our setting, we train it on the FullBodyManipulation dataset without modifying its model structure.

\begin{table*}
  \centering
  \begin{tabular}{cccccccccccc} 
    \toprule
     \multicolumn{3}{c}{Components}    & \multicolumn{3}{c}{R-precision$\uparrow$}  & \multirow{2}{*}{FID$\downarrow$}  & \multirow{2}{*}{Diversity$\rightarrow$} & \multirow{2}{*}{FS$\downarrow$} & \multirow{2}{*}{$C_{\%}$} & \multirow{2}{*}{$Coll_{\%}$} & \multirow{2}{*}{MDev$\downarrow$}\\  
     
    \cmidrule(r){1-3} \cmidrule(r){4-6}
    obj-kp & IDF loss & ~~$G$~~  & Top-1 & Top-2 & Top-3 & & & & \\
    \midrule
     {$\times$} &  {$\times$}&  {$\times$}  & 0.495 & 0.626 & 0.681 & 9.775  & 9.972& \textbf{0.331} & 0.349 & 0.210 & 9.549\\
    {\checkmark} & {$\times$}&  {$\times$} & 0.547  &0.686  & 0.767 & 7.514  & 
     9.738 & 0.417 & 0.374 & 0.205 & 9.227\\
    {\checkmark} & {\checkmark} &  {$\times$}   & 0.666 & 0.817 & 0.879 & 5.726  & 8.915 & 0.415 & 0.424 & 0.194 & 7.020\\
    {\checkmark} & {\checkmark} &  {$C$}  & 0.668 & 0.839 & 0.892 & 5.902  & 8.910 & 0.371 & 0.364 & \textbf{0.176} & 9.936\\
    {\checkmark} & {\checkmark} & {$D$}   & \textbf{0.706}   & \textbf{0.835}   & \textbf{0.902} & \textbf{5.119}  & \textbf{8.863}& 0.349 & \textbf{0.466} & 0.200 & \textbf{5.815} \\
    \bottomrule
  \end{tabular}
  \caption{Ablation study on the FullBodyManipulation dataset \cite{li2023object}. Starting with the baseline, we incrementally incorporate three components to evaluate their individual impact on HOI synthesis. `C' denotes using a distance matrix that contains only the object centroid and human joints, while `D' represents using a full distance matrix that includes both object key points and human joints (our proposed setting).}
  \label{sample-table2}
\end{table*}

\subsection{Quantitative Results}

Table~\ref{sample-table1} presents our quantitative results on the FullBodyManipulation dataset, comparing \modelname with four baseline models. Our approach demonstrates significant improvements across many metrics. Notably, \modelname achieves a substantial increase in R-Precision, indicating better alignment between generated motions and textual descriptions. Compared to CHOIS, the most advanced baseline, our method boosts Contact Percentage without increasing Collision Percentage, suggesting more physically plausible contacts. Furthermore, the increase in the MDev metric reflects improved synchronization and consistency between human and object movements. These results underscore the effectiveness of our relational modeling approach in capturing complex spatiotemporal relationships, guiding generated motions, and producing more realistic and coherent human-object interactions.

\subsection{Qualitative Results}

We visualize the generated HOIs from each baseline, with the results presented in Figure~\ref{fig:compare}. Both InterGen and MDM exhibit unnatural object movements and noticeable gaps between human and object contacts, significantly reducing the realism of the interactions. While HOI-Diff \cite{peng2023hoi} and CHOIS \cite{li2023controllable} employ guidance mechanisms to direct human-object interactions, each has inherent limitations. HOI-Diff relies on fixed contact points between human and objects, performing well in scenarios with constant contact but struggling in dynamic, multi-stage interactions such as the example in Figure~\ref{fig:compare}, where the human initially kicks a trashcan and later picks it up and places it. This rigidity results in jitter and unnatural motion. Similarly, CHOIS imposes guidance exclusively on hand-object interactions, neglecting contact from other body parts, and leading to incomplete modeling of human-object dynamics. In contrast, our model excels at generating coherent and natural interactions that capture the evolving relationship between human and objects. This adaptability allows our model to produce more realistic and contextually appropriate interactions, demonstrating its effectiveness.

\subsection{Ablation Studies}
\begin{figure}[t]
  \centering
  \includegraphics[width=0.95\linewidth,height=8.5cm]{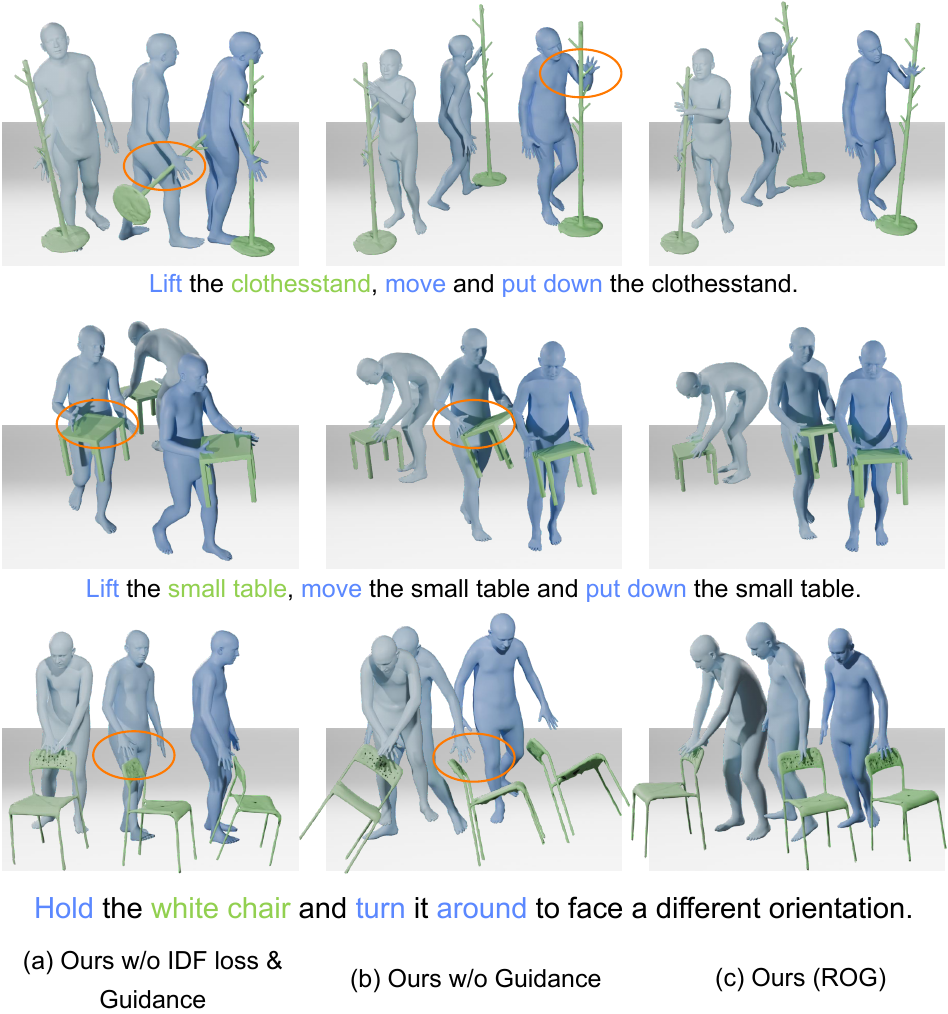}
    \caption{Ablation visual results. We systematically assemble the model, starting from a basic baseline and incrementally adding our innovative components.}
    \label{fig:ablation}
\end{figure}

To assess the contribution of each component in our proposed \modelname, we conduct a series of ablation experiments by incrementally adding components to evaluate their individual impacts. The results are summarized in Table~\ref{sample-table2} and visualized in Figure~\ref{fig:ablation}.

Initially, incorporating object key points (\textbf{obj-kp}) into the input provides a more comprehensive representation of objects, thereby enhancing the realism of motion, as demonstrated by improved FID score. Building on this, the introduction of the distance loss function (\textbf{IDF loss}) further optimizes the spatial relationships in human-object interactions, resulting in better coordination. This improvement is reflected in higher contact percentages, lower collision rates, and improved MDev scores, while also enhancing the alignment between generated motions and text prompts, as indicated by higher R-Precision scores. Moreover, using the full distance matrix ($D$) significantly outperforms both the centroid distance matrix ($C$) and the absence of guidance, with superiority evident across metrics such as semantic alignment (R-Precision), realism (FID), and others. These results collectively underscore the effectiveness of integrating each component in enhancing performance across multiple dimensions.

\section{Conclusion and Limitations}

In this work, we introduce \textbf{\modelname}, a novel diffusion-based framework designed to model the spatiotemporal relationships inherent in human-object interactions with detailed geometric precision. We begin by constructing an Interactive Distance Field (IDF) that captures the dynamics of HOIs by leveraging both boundary-focused and fine-detail key points on the object mesh, ensuring an accurate representation of the object's geometry. Building on this, we develop a diffusion-based relation model to refine the IDF of the generated motion, guiding the motion generation process to produce relation-aware and semantically aligned movements.
Limitations include dataset constraints: excluded hand/finger motions result in low-fidelity hand details.

\paragraph{Broader Impacts.} Human-object interaction synthesis enhances applications such as virtual reality, computer animation, and robotics by enabling more immersive and realistic experiences. To the best of our knowledge, this work has no obvious negative social impacts.

\paragraph{Acknowledgement.} This work was supported by the National Natural Science Foundation of China under Grant 62476099 and 62076101, Guangdong Basic and Applied Basic Research Foundation under Grant 2024B1515020082 and 2023A1515010007, the Guangdong Provincial Key Laboratory of Human Digital Twin under Grant 2022B1212010004, the TCL Young Scholars Program, and the 2024 Tencent AI Lab Rhino-Bird Focused Research Program.


\clearpage
\setcounter{page}{1}
\setcounter{section}{0}
\maketitlesupplementary

\section{Overview}
This supplementary material provides detailed insights into our method, covering evaluation details (Section~\ref{Evaluation Details}), extended experiments on guidance mechanisms, T2M-BEHAVE Dataset, and novelty assessment (Section~\ref{Experiments}), performance comparisons with baselines (Section~\ref{Performance_comparison}), user study results (Section~\ref{userstudy}), and qualitative results demonstrating the model's robustness (Section~\ref{Qualitative Results}). Additionally, we include comprehensive details on motion representation and architecture for clarity and reproducibility (Section~\ref{methoddetails}), along with a discussion on hand motion synthesis and future extensions (Section~\ref{hand motion}).

\section{Evaluation Details}
\label{Evaluation Details}
    \paragraph{Text and HOI Feature Extraction.} 
Currently, there are no publicly available feature extractors specifically designed to evaluate human-object interaction motions. To address this limitation, we take inspiration from T2M \cite{guochuan}  and adopt a similar evaluation framework. Our method converts the textual descriptions into feature vectors with a frozen CLIP text encoder. At the same time, the generated HOI sequences are processed using an HOI feature extractor based on a bidirectional GRU (BiGRU) model. Moreover, we modify the input dimensions of the BiGRU model according to the representation of HOI sequences. Specifically, the length of the representation is 216, with 72 dimensions allocated to the human’s 24 joints, 132 dimensions representing 6D continuous rotations, 9 dimensions for the object’s rotation matrix, and 3 dimensions for object transformations. By minimizing the distance between features of the matched text-HOI pairs, this approach establishes a strong alignment between the textual descriptions and HOI motion sequences.

    \paragraph{Contact Percentage.} 
Following the method in CHOIS \cite{licontrollable}, we compute the Contact Percentage by calculating the minimum distance between hand joints and object vertices at each frame. A 5 cm threshold is used to determine contact. The Contact Percentage is then the ratio of frames with contact to the total number of frames, representing the proportion of time during which hand-object contact occurs.
    
    \paragraph{Collision Percentage.} 
In line with OMOMO \cite{liobject}, we compute the Collision Percentage by querying the object's Signed Distance Function (SDF) at each time step for each vertex on the reconstructed human mesh. A 4 cm threshold is applied to detect collisions, where a collision is counted if the signed distance is negative and its absolute value exceeds 4 cm. The Collision Percentage is calculated as the ratio of frames with collisions to the total number of frames in the sequence.

    \paragraph{Motion Deviation.}
Following the approach in ARCTIC \cite{fanarctic}, we define the Motion Deviation (MDev) metric to assess the consistency of motion between the hand and object vertices in a HOI sequence. MDev measures the difference in movement directions between the hand and object vertices over consecutive frames within a window \( (m, n) \). Let \( h_i^t \) and \( o_j^t \) represent the hand and object vertices at frame \( t \), respectively. A contact window \( (i, j, m, n) \) is defined as the longest period during which \( h_i^t \) and \( o_j^t \) remain within a threshold \( \alpha = 5 \, \text{cm} \) for all frames between \( m \) and \( n \), with no contact at frames \( m-1 \) and \( n+1 \). The MDev is then calculated as:
\[
\text{MDev} = \frac{1}{n-m} \sum_{t=m+1}^n \left\| \left(\hat{\mathbf{h}}_i^t - \hat{\mathbf{h}}_i^{t-1}\right) - \left(\hat{\mathbf{o}}_j^t - \hat{\mathbf{o}}_j^{t-1}\right) \right\|,
\]
where MDev quantifies the movement consistency between the contacting hand and object vertices. The final MDev value is the average of MDev values across all detected windows, with the result expressed in millimeters (mm).

\section{Additional Experiments}
\label{Experiments}

    \paragraph{Impact of Guidance Mechanism at Different Noise Levels.}
In this section, we examine the impact of the diffusion timestep selection on the effectiveness of IDF Guidance. To assess how different timestep settings affect generation performance, we conducted a series of experiments. As summarized in Table~\ref{guidance_impact}, the results show that initiating IDF Guidance during the final 10 timesteps yields the best results. This approach outperforms both applying IDF Guidance throughout all 1000 timesteps and applying it solely during the final timestep, highlighting the importance of timing the guidance effectively for optimal results.

\begin{table}
  \centering
  \setcounter{table}{0}
  \begin{tabular}{lcccc}
    \toprule
    Methods & Precision$\uparrow$ & FID$\downarrow$ & $C_{\%}$ & MDev$\downarrow$ \\ 
    \midrule
    {w/o Guidance} & 0.879 & 5.726 & 0.424 & 7.020 \\
    \cmidrule(lr){1-5}
    {\( t \leq 0.001T \)} & 0.888 & 5.159 & 0.458 & 6.435 \\
    {\( t \leq 0.005T \)} & 0.896 & 5.129 & 0.463 & 5.927\\
    {\( t \leq 0.01T \)}  & \textbf{0.902} & \textbf{5.119} & \textbf{0.466} & \textbf{5.815} \\
    {\( t \leq 0.1T \)} & 0.890 & 5.399 & 0.464 & 6.590 \\
    {\( t \leq 0.5T \)} &0.879 & 5.586 & 0.430 & 6.938 \\
    {\( t \leq T \)} &0.879 & 5.590 & 0.431 & 6.954 \\
    \bottomrule
  \end{tabular}
    \caption{Quantitative results showing the impact of guidance timesteps on generation performance during diffusion. Initiating IDF Guidance during the final 10 timesteps (\(t \leq 0.01T\)) consistently achieves the best performance across metrics. Precision refers to the R-precision top-3 metric.}
  \label{guidance_impact}
\end{table}

    \paragraph{Effect of Guidance Iterations.}
This section examines the influence of the number of guidance iterations on the performance of IDF-guided denoising. Specifically, we perform \( k \) iterations of L-BFGS optimization at each denoising step, where \( k \) is treated as a tunable hyperparameter. The experiments are conducted during the last 10 timesteps of the denoising process to refine the generation results. 
As summarized in Table~\ref{guidance_iterations}, increasing \( k \) to 10 demonstrates the most significant improvement, achieving the best balance across metrics such as Precision, FID, and Motion Deviation. This outcome highlights the importance of iterative refinement in aligning distributions, as second-order optimization enables more precise adjustments and accelerates convergence towards the target solution.

\begin{table}
  \centering
  \begin{tabular}{lcccc}
    \toprule
    Methods & Precision$\uparrow$ & FID$\downarrow$ & $C_{\%}$ & MDev$\downarrow$ \\ 
    \midrule
    {w/o Guidance} & 0.879 & 5.726 & 0.424 & 7.020 \\
    \cmidrule(lr){1-5}
    {\( k = 1 \)} & 0.883 & 5.597 & 0.430 & 6.837 \\
    {\( k = 5 \)} & 0.900 & 5.347 &  \textbf{0.474} & 6.482\\
    {\( k = 10 \)}  & \textbf{0.902} & \textbf{5.119} & 0.466 & \textbf{5.815} \\
    \bottomrule
  \end{tabular}
    \caption{Effect of guidance iterations (\(k\)) on generation performance during the final 10 denoising steps. Increasing \(k\) enhances performance, with \(k = 10\) achieving the best results.}
  \label{guidance_iterations}
\end{table}

    \paragraph{Influence of the Relation Model.}
We evaluated the impact of the Relation Model by replacing the original VDT model \cite{vdt} with the MDM model \cite{mdm} and increasing the number of Transformer encoder layers in MDM to expand its parameters during training. The results, summarized in Table~\ref{relation_model_comparison}, indicate that VDT consistently outperforms MDM across all metrics. This advantage is attributed to VDT's spatial and temporal self-attention mechanisms, which enable it to effectively capture complex dependencies, leading to higher-quality outputs.

\begin{table}
  \centering
  \begin{tabular}{lccccc}
    \toprule
    $R\left(\cdot \right)$ &Params &Precision$\uparrow$ & FID$\downarrow$ & $C_{\%}$ &MDev$\downarrow$ \\ 
    \midrule
    {MDM} & 28.79M & 0.887& 5.529& 0.370& 6.204 \\
    {MDM} &30.89M &0.890 &5.358 &0.431 &7.401  \\
    {MDM} &33.00M &0.896 &5.384 &0.410 &6.703  \\
    {Ours} & 29.36M & \textbf{0.902} & \textbf{5.119} & \textbf{0.466} & \textbf{5.815} \\
    \bottomrule
  \end{tabular}
    \caption{Performance comparison of relation models $R\left(\cdot \right)$. Ours, based on the VDT model with spatial and temporal attention mechanisms, achieves superior precision, FID, and MDev metrics compared to the MDM model, despite having comparable or smaller parameter sizes.}
  \label{relation_model_comparison}
\end{table}

    \paragraph{Results on T2M-BEHAVE Dataset.}
\begin{table}
  \centering
  \footnotesize 
  \resizebox{0.96\linewidth}{!}{ 
    \begin{tabular}{lcccc}
      \toprule
      Method & FS $\downarrow$ & $C_{\%}$ & $Coll_{\%}$ & MDev$\downarrow$ \\ 
      \midrule
      {GT}  & 0.071 & 0.410 & 0.370 & 8.989 \\
      {HOI-Diff\cite{Hoi-diff}}  & 0.182 & 0.191 & 0.259 & 24.807 \\
      {Ours}  & \textbf{0.174} & \textbf{0.239} & \textbf{0.195} & \textbf{10.784}\\
      \bottomrule
    \end{tabular}
  }
  \caption{T2M-BEHAVE \cite{Hoi-diff} cross-benchmark tests: 24.6\% lower collisions vs HOI-Diff (0.195 vs 0.259), despite dataset's compact scale.}
  \label{behave_dataset_comparison}
\end{table}
As FullBodyManipulation (FBM) is much larger than other text-conditioned HOI datasets, both our work and CHOIS \cite{licontrollable} adopt FBM only for benchmarking. We have included evaluations on T2M-BEHAVE \cite{Hoi-diff} in Table~\ref{behave_dataset_comparison}. The results highlight inherent dataset characteristics, with ground-truth (GT) collision metrics registering at 0.370.
While the dataset's compact scale and lack of dedicated interaction-focused evaluation protocols limit comprehensive benchmarking, our method demonstrates improved physical plausibility by reducing collisions to 0.195 compared to HOI-Diff's 0.259. The motion quality metrics (FS: 0.174 vs. 0.182; MDev: 10.784 vs. 24.807) indicate our approach maintains reasonable fidelity.

    \paragraph{Retrieval vs Generation}
\begin{figure}[t]
  \centering
    \setcounter{figure}{0}
  \includegraphics[width=0.8\linewidth]{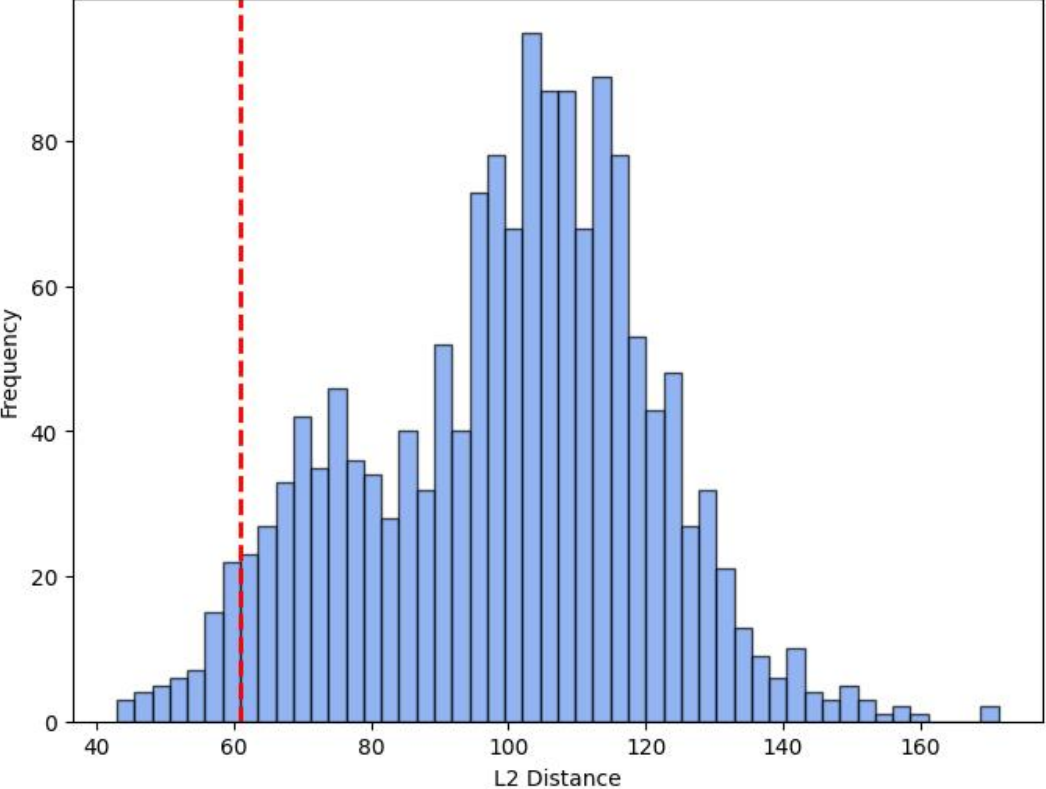}
    \caption{Novelty analysis. L2 distances of 512 test cases vs. top-3 training neighbors. Red lines mark intra-trainset distance, demonstrating generation beyond training data.}
    \label{fig:NoveltyAnalysis}
\end{figure}
Following CG-HOI \cite{Cg-hoi}, Figure \ref{fig:NoveltyAnalysis} first detects top-3 nearest training samples for each of 512 test cases, then plots an L2 distance histogram. It also marks the intra-trainciteset distance with a red line. It shows the generated motions mostly fall outside the intra-trainset distance, which confirms our model produces novel motions rather than retrieval.

\section{Model Complexity and Inference Efficiency}
\label{Performance_comparison}
\begin{table}
  \centering
    \resizebox{0.96\linewidth}{!}{ 
  \begin{tabular}{lccccc}
    \toprule
    Metric & InterGen & MDM & HOI-Diff & CHOIS & Ours \\ 
    \midrule
    Params(M) & 54.24 & 17.91 & 47.74 & 13.25 & 47.34 \\
    Inference time(s) & 2.0 & 0.36 & 3.68 & 2.93 & 8.1 \\
    \bottomrule
  \end{tabular}}
  \caption{Comparison of model complexity and inference efficiency.}
  \label{performance_comparison}
\end{table}

Table~\ref{performance_comparison} compares computational costs for 10-second motion generation on an RTX 4090 GPU, with all models trained for 300,000 steps. Our method combines a motion diffusion model (17.98M parameters) and a relation diffusion model (29.36M parameters). With a batch size of 32, our method achieves an average of 8.1 seconds per sequence, remaining viable compared to baselines (MDM: 0.36s, HOI-Diff: 3.68s).

\section{User study}
\label{userstudy}
\begin{figure}[t]
  \centering
  \includegraphics[width=0.9\linewidth]{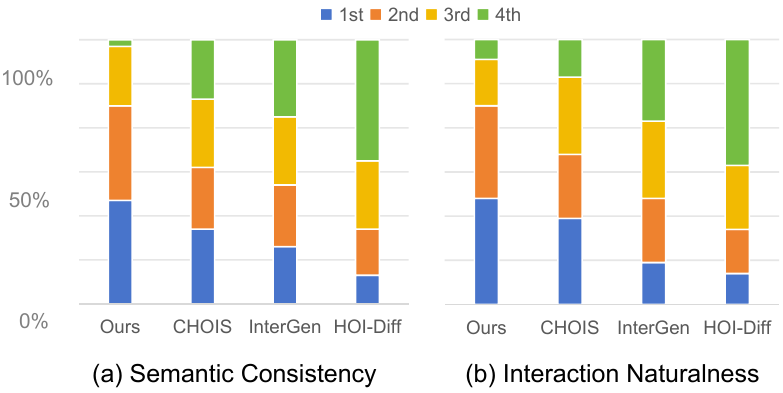}
    \caption{User study. We generated HOIs for 15 captions using 4 methods and asked 20 users to rank them by text alignment and realism. Our method outperforms others in both aspects.}
    \label{fig:Userstudy}
\end{figure}

We conducted a user study comparing four HOI generation methods across 15 text captions, with twenty participants ranking the results based on two evaluation criteria: (1) Semantic Consistency (alignment between animations and text descriptions) and (2) Interaction Naturalness (naturalness of poses and object interactions). As shown in Figure \ref{fig:Userstudy}, our method outperformed three baseline methods in both metrics, demonstrating superior text-visual correspondence and biomechanical plausibility.

\begin{figure*}[t]
  \centering
  \includegraphics[width=1.0\linewidth]{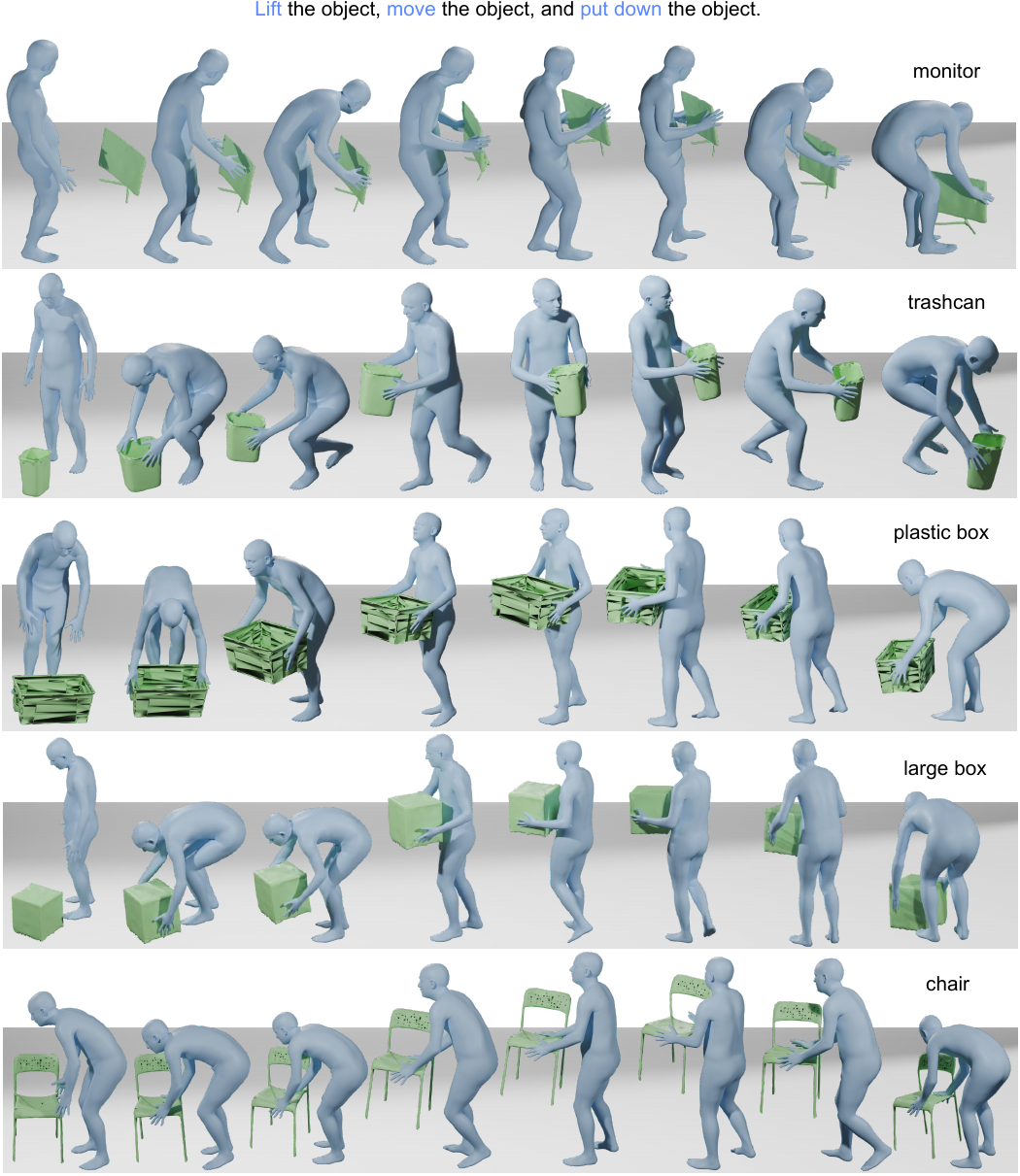}
    \caption{Our model generates semantically accurate and consistent human-object interactions across various objects.}
    \label{fig:vis1}
\end{figure*}

\begin{figure*}[t]
  \centering
  \includegraphics[width=1.0\linewidth]{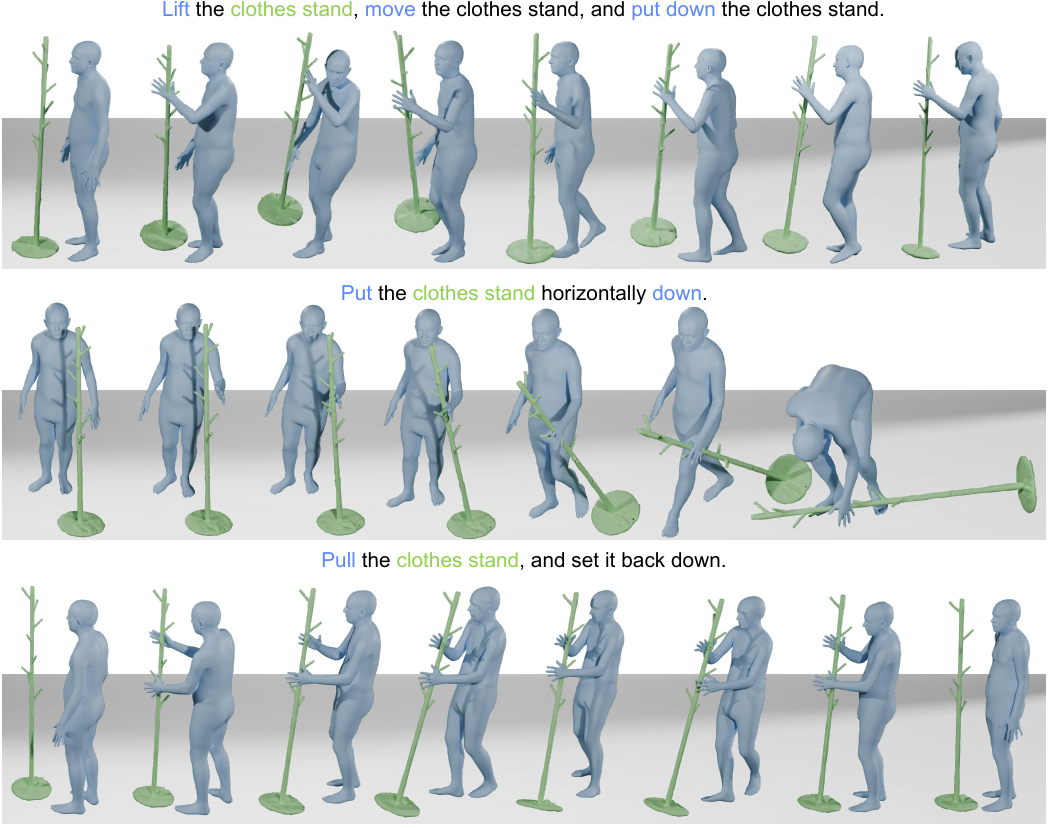}
  \caption{Our model generates diverse and semantically distinct human-object interactions for a single object.}
  \label{fig:vis2}
\end{figure*}

\section{Additional Qualitative Results}
\label{Qualitative Results}
In this section, we provide additional HOI generation results across diverse settings, highlighting the versatility and effectiveness of our approach.

\paragraph{Consistent Motions across Different Objects.}
We assess our method's ability to generate consistent motions across different scenarios, such as ``lift the object, move the object, and put down the object" across different objects. Figure \ref{fig:vis1} illustrates the interaction results for a monitor, trashcan, plastic box, large box, and chair. These examples highlight ROG's ability to produce semantically accurate and consistent interactions across a range of objects.

    \paragraph{Diverse Motions on the Same Object.}
We evaluate the ability of \modelname to generate diverse motions for the same object. Taking a clothes stand as an example, Figure~\ref{fig:vis2} showcases a range of motions, including ``lift the object", ``put down the object", and ``pull the object". These results demonstrate the flexibility of our approach in producing semantically distinct and contextually appropriate interactions with a single object.

\section{Method Details}
\label{methoddetails}
\paragraph{Motion representation Details.} 
The motion data is represented in a 288-dimensional vector per frame, comprising two primary components: human motion and object-related information. The 204-dimensional human motion data captures both the global 3D coordinates of 24 body joints (24×3) and the 6D rotational parameters for 22 joints (excluding palm joints, 22×6). The remaining 84 dimensions describe object interactions through three key elements: the object's global position (3), its 3×3 rotation matrix (9), and the spatial coordinates of 24 predefined key points on the object's surface (24×3).

\paragraph{Architecture Details.}
\begin{figure*}[t]
  \centering
  \includegraphics[width=1.0\linewidth]{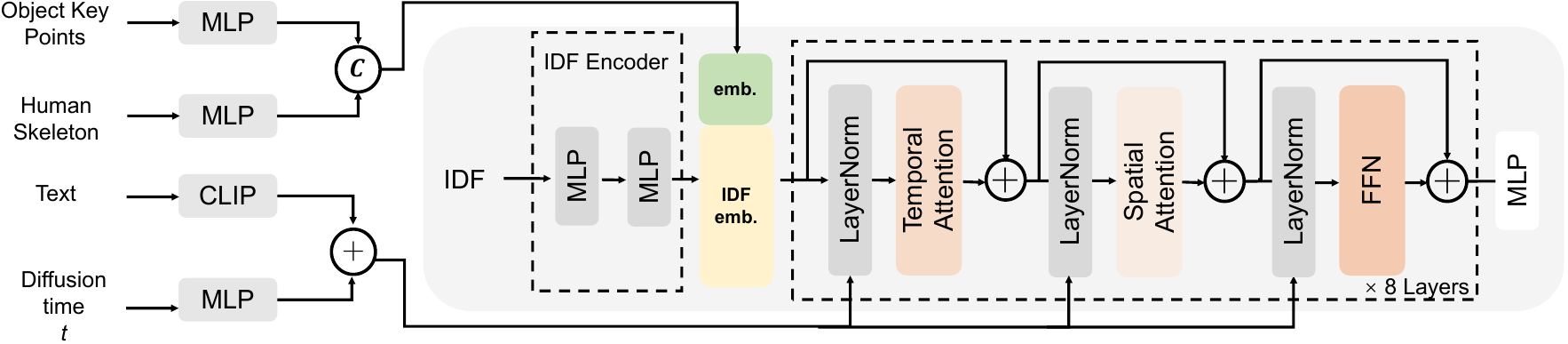}
    \caption{Overview to our Relation Model’s architecture. It consists of an IDF encoder and multiple VDT layers.}

    \label{fig:architecture}
\end{figure*}
Figure~\ref{fig:architecture} illustrates the details of our relation model, including the IDF encoder and VDT layers.

\section{Hand Motion Synthesis and Future Extensions}
\label{hand motion}
While our framework currently omits fine-grained hand motion synthesis, this limitation primarily stems from the lack of comprehensive hand-object interaction data in mainstream HOI benchmarks such as FullBodyManipulation and BEHAVE. Existing datasets predominantly focus on coarse full-body dynamics, leaving finger-level articulations under-explored. By extending the Interactive Distance Field (IDF) to incorporate hand-specific keypoints (e.g., fingertip positions), our framework can seamlessly adapt to richer representations. This flexibility demonstrates the broad applicability of our method across different motion scales, enabling the generation of both body movements and hand manipulations when trained on appropriate data.

\end{document}